%% file: main.tex
\documentclass[lettersize,journal]{IEEEtran}
\usepackage{amsmath,amsfonts}
\usepackage{algorithmic}
\usepackage{algorithm}
\usepackage{array}
\usepackage[caption=false,font=normalsize,labelfont=sf,textfont=sf]{subfig}
\usepackage{textcomp}
\usepackage{stfloats}
\usepackage{url}
\usepackage{verbatim}
\usepackage{graphicx}
\usepackage{cite}
\usepackage{multirow}
\usepackage{booktabs}
\usepackage{enumitem}
\usepackage[colorlinks, linktoc=all]{hyperref}
\hypersetup{allcolors=blue}
\usepackage{cleveref}
\crefname{figure}{Fig.}{Figs.}
\crefname{table}{Table}{Tables}
\crefname{algorithm}{Algorithm}{Algorithms}
\crefname{section}{Section}{Sections}

\begin{document}

\title{$\Lambda$-Split: A Privacy-Preserving Split Computing Framework for Cloud-Powered Generative AI}
\author{
Shoki~Ohta and 
Takayuki~Nishio
\thanks{S. Ohta and T. Nishio are with the School of Engineering, Tokyo Institute of Technology, Tokyo 152-8550, Japan (e-mail: \href{mailto:nishio@ict.e.titech.ac.jp}{nishio@ict.e.titech.ac.jp}).}
}

\markboth{Journal of \LaTeX\ Class Files,~Vol.~14, No.~8, August~2021}%
{Shell \MakeLowercase{\textit{et al.}}: A Sample Article Using IEEEtran.cls for IEEE Journals}


\maketitle

\begin{abstract}
In the wake of the burgeoning expansion of generative artificial intelligence (AI) services, the computational demands inherent to these technologies frequently necessitate cloud-based computational offloading, particularly for resource-constrained mobile devices. 
These services commonly employ prompts to steer the generative process, and both the prompts and the resultant content—such as text and images—may harbor privacy-sensitive or confidential information, thereby elevating security and privacy risks. 
To mitigate these concerns, we introduce $\Lambda$-Split, a split computing framework to facilitate computational offloading while simultaneously fortifying data privacy against risks such as eavesdropping and unauthorized access.
The $\Lambda$-Split framework is conceptually akin to federated learning, aiming to preserve data privacy by confining privacy-sensitive data to local user devices. 
Specifically, in $\Lambda$-Split, a generative model—usually a deep neural network (DNN)—is partitioned into three sub-models and distributed across the user's local device and a cloud server: the input-side and output-side sub-models are allocated to the local, while the intermediate, computationally-intensive sub-model resides on the cloud server. 
This architecture ensures that only the hidden layer outputs are transmitted, thereby preventing the external transmission of privacy-sensitive raw input and output data. 
Given the black-box nature of DNNs, inferring the original input or output from intercepted hidden layer outputs poses a significant challenge for malicious eavesdroppers. 
Moreover, $\Lambda$-Split is orthogonal to traditional encryption-based security mechanisms, offering enhanced security when deployed in conjunction. 
We empirically validate the efficacy of the $\Lambda$-Split framework using Llama 2 and Stable Diffusion XL—representative large language and diffusion models developed by Meta and Stability AI, respectively. 
We conclude by outlining future research avenues in the realm of privacy-preserving distributed AI over wireless networks. 
Our $\Lambda$-Split implementation is publicly accessible at \url{https://github.com/nishio-laboratory/lambda_split}.
\end{abstract}


\section{Introduction}

The remarkable strides in generative artificial intelligence (AI) technologies, notably represented by large language models (LLMs) and diffusion models, have facilitated the generation of text and images that are virtually indistinguishable from human-produced content~\cite{zhao2023llmsurvey, yang2023dmsurvey}. 
Generative AI services, such as ChatGPT\footnote{\url{https://chat.openai.com} (accessed on October 15, 2023)} and DALL·E, have not only emerged but are also swiftly gaining widespread adoption.
A myriad of individuals and enterprises are eager to capitalize on these innovations for a wide array of applications, including automated responses, software code generation, and automated multimedia content creation.
These generative processes often compel the offloading of computational tasks to cloud computing infrastructures, particularly for resource-constrained devices such as laptops and smartphones, due to the computational intensity of generative models.
The generative instructions, colloquially known as ``prompts," along with the generated text and images, are relayed between user devices and cloud servers through communication networks such as wireless communication channels and the Internet. 
Nonetheless, these prompts and the resultant generated content may contain privacy-sensitive or confidential information, thereby introducing inherent risks of eavesdropping and data leakage during wireless transmission.
To realize sustainable generative AI services, privacy preservation is an indispensable factor.

\begin{figure*}[t!]
    \centering
    \includegraphics[width=\linewidth]{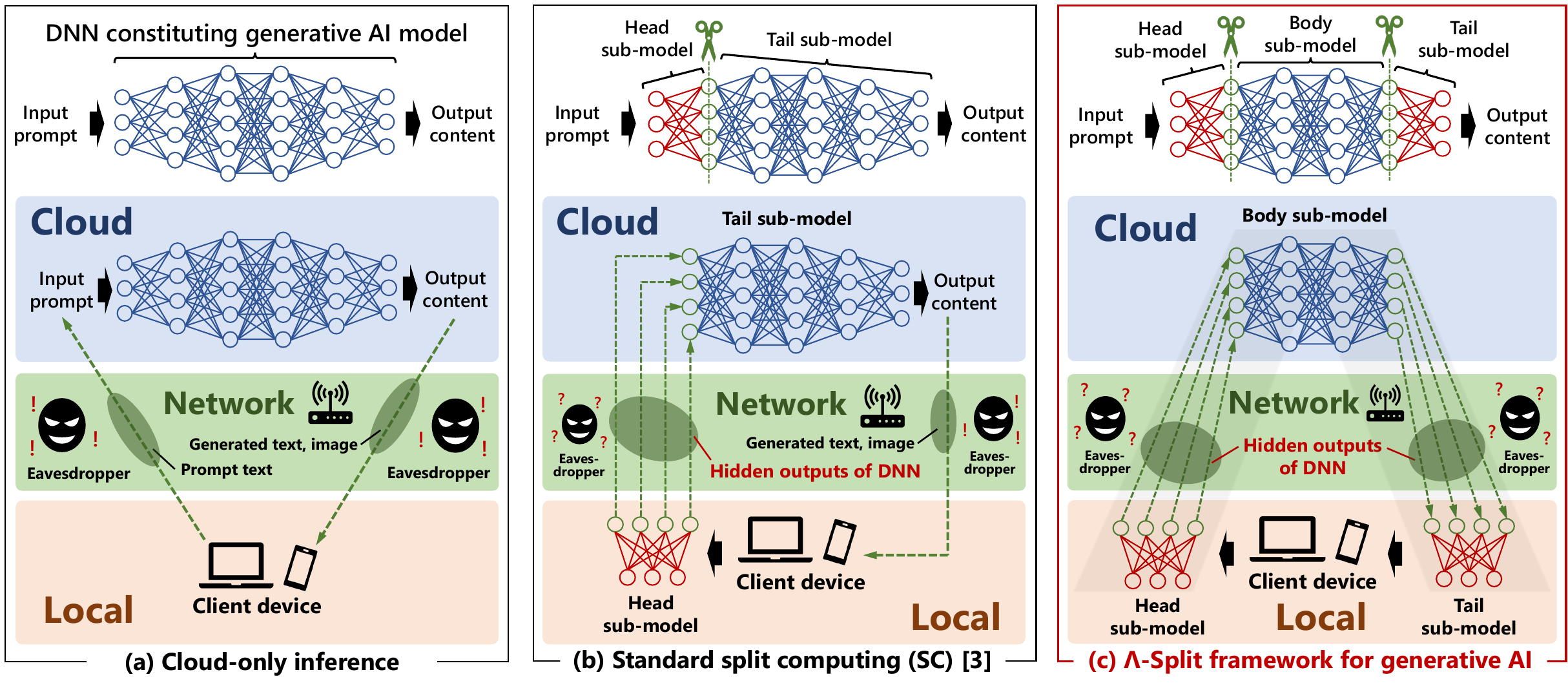}
    \caption{Schematic representation of a generative model implemented using the cloud-only inference, standard SC~\cite{matsubara:sc_survey}, and $\Lambda$-Split framework. 
    $\Lambda$-Split enhances privacy preservation because all transmitted data is the hidden output of DNN while leveraging the computational power of the cloud.
    The nomenclature of $\Lambda$-Split is emblematic of this data flow pattern, where the local-to-cloud-to-local fold-back trajectory resembles the Greek letter $\Lambda$. 
    }
    \label{fig:system}
\end{figure*}

To mitigate privacy concerns associated with cloud-powered generative AI, we unveil a novel framework, $\Lambda$-Split, which employs split computing (SC) to simultaneously bolster data privacy and facilitate computational offloading. 
A schematic overview is delineated in \cref{fig:system}. 
SC was originally formulated to distribute the computational workload of deep neural networks (DNNs) across multiple devices, typically involving a local device and a cloud server~\cite{matsubara:sc_survey}, and serves as the architectural cornerstone for our approach. 
Within the standard SC paradigm, a DNN is bifurcated into input-side and output-side sub-models, which are judiciously allocated to a local device and a cloud server, respectively. 
The inference phase is executed by relaying the hidden layer outputs from the local device's input-side sub-model to the cloud server. 
Exploiting the unique feature where only hidden layer outputs are transmitted, we introduce $\Lambda$-Split, a specialized split computing framework that emphasizes privacy preservation through a tripartite division of the DNN. 
While $\Lambda$-Split shares conceptual similarities with federated learning (FL)~\cite{itahara_TMC} in its aim to retain privacy-sensitive data within local user devices, it diverges by extending this privacy-preserving principle to the inference process. 
Specifically, $\Lambda$-Split is designed to ensure that neither the model's input nor its output data are externalized from the local device.

In $\Lambda$-Split framework, a generative model—usually a DNN—is partitioned into three sub-models: the initial (input-side) and terminal (output-side) sub-models are allocated to the user’s local devices, while the intermediate, computationally-intensive sub-model resides on a cloud server. Hence, both input and output data no longer traverse the communication network while leveraging the computational capabilities of the cloud server.
Even if eavesdroppers capture the hidden layer vector, eavesdroppers are difficult to decode the original input or output data owing to the black-box characteristic of DNNs. 
While this framework does not guarantee computational security in the same manner as encryption technologies, it introduces a novel approach that is orthogonal to encryption, achieving both privacy preservation and computation offloading simultaneously. 
Consequently, $\Lambda$-Split holds the potential for further enhancing overall security when used in conjunction with encryption technologies.

In this article, we present specific application methodologies for $\Lambda$-Split in the context of text generation using state-of-the-art LLMs and image generation employing diffusion models. 
We propose model splitting strategies and communication traffic reduction methods, and experimentally demonstrate the feasibility of $\Lambda$-Split for LLM and diffusion model. 
Additionally, we engage in a discussion of future research directions. 
The contributions are summarized as follows:

\begin{itemize}
    \item We introduce $\Lambda$-Split as a groundbreaking framework for privacy preservation in cloud-powered generative models. 
    This framework is orthogonal to traditional encryption methods and can be synergistically employed alongside them. 
    $\Lambda$-Split extends the privacy-preserving principles of FL, which aims to keep privacy-sensitive data localized to user devices, to the inference phase of model operation.

    \item We have actualized a $\Lambda$-Split-based distributed generative AI system for two state-of-the-art generative models: Llama 2~\cite{touvron2023llama2}, a LLM for text generation developed by Meta, and Stable Diffusion XL (SDXL)~\cite{podell2023sdxl}, a diffusion model for image generation developed by Stability AI. 
    We have implemented a web application equipped with a user interface for $\Lambda$-Split utilizing HTTP and HTML, and visualized eavesdropping tolerance.
    The source code is publicly accessible at \url{https://github.com/nishio-laboratory/lambda_split}.

    \item We incorporate traffic reduction strategies that utilize a caching mechanism for the LLM and a quantization technique for the diffusion model. 
    The efficacy of these methods has been empirically validated.
    
    \item We discuss challenges and potential research opportunities toward further sustainable generative AI associated with $\Lambda$-Split.
\end{itemize}

\section{System model and general framework}
The generative AI system deployed under the $\Lambda$-Split framework consists of local user devices, including but not limited to smartphones, laptops, and embedded computing systems, in conjunction with a computationally high-capacity cloud server. 
While these local devices may lack the computational power to process the entire generative model, they are capable of handling a split sub-model of the model. 
These local devices are connected to the cloud server via communication networks. 
We operate under the assumption that eavesdroppers exist within these networks, actively intercepting packets to illicitly obtain private or confidential information from generative AI users. 
Additionally, we assume that pre-trained generative models are available. 

We delineate the overarching application workflow of the $\Lambda$-Split framework, which is fundamentally segmented into two primary phases: the \texttt{preparation phase} and the \texttt{inference phase}. 
During the \texttt{preparation phase}, the cloud server initially identifies the optimal splitting points to divide the generative model into three distinct sub-models. 
These are categorized as the input-side, middle, and output-side sub-models, colloquially termed the head, body, and tail sub-models, respectively.
The selection of these splitting points must be judiciously made, considering factors such as the computational capacity of the local device, the bandwidth of the communication link, and the desired level of privacy protection. 
Generally, a greater number of parameters leads to increased computational latency; therefore, it is preferable for the head and tail sub-models deployed on the local device to have fewer layers. 
However, a trade-off arises when the splitting occurs too close to the input or output, as this reduces the number of mappings to the input or output data, thereby increasing the risk of information leakage. 
Additionally, the dimensionality of the hidden layer outputs at the splitting points determines the network traffic generated during inference. 
Splitting at layers with a large number of nodes or channels, or in the middle of skip-connections, can result in substantial communication overhead. 
Consequently, the optimal splitting points must be carefully selected based on the specific model and performance requirements.

Subsequently, customization of the model is often undertaken; although not mandatory, this step frequently involves various optimizations, particularly for the head sub-model executed on the local device. 
Customization of the model for $\Lambda$-Split remains an open issue, which will be discussed as a research direction in the "Challenges and Potential Research Opportunities" section. 
Concurrently, the customized head and tail sub-models are deployed to the local device, a process that necessitates the implementation of secure communication protocols to thwart potential compromise by malicious actors.

In the \texttt{inference phase}, the local device initially feeds the prompt into the head sub-model and forwards the resultant output vector to the cloud server. 
Subsequently, the cloud server processes this output through the body sub-model, generating an output vector that is then transmitted back to the local device. Finally, the tail sub-model on the local device processes this vector to produce the final generated content, such as text or images. 
Throughout this inference pipeline, only the hidden layer outputs of the DNN traverse the communication network, ensuring that neither input prompts nor generated contents are exposed beyond the local device. 
Given the non-linearity and black-box characteristics of DNNs, the reconstruction of input or output data based solely on these hidden outputs presents a formidable challenge. 
Consequently, our framework amplifies resistance to eavesdropping and data leakage, thereby enhancing privacy preservation.

As underscored in the exposition of the \texttt{preparation phase}, a pivotal challenge in the $\Lambda$-Split framework lies in the astute splitting of the model. 
Modern generative AI models are far from being simple sequential architectures; they encompass intricate mechanisms such as branching, autoregression, and loops. 
Without comprehensive knowledge of the model's architecture, the endeavor to effectively partition the inference pipeline becomes complex, potentially leading to a significant increase in the volume of transmitted data. 
In the ensuing sections of this article, we showcase two exemplary applications of $\Lambda$-Split: the first focuses on text-to-text generation using LLMs, and the second addresses text-to-image generation through diffusion models. 
Additionally, we propose tailored strategies for reducing communication traffic in alignment with the architecture of each respective model.

\section{$\Lambda$-Split for Large Language Model-based Text Generation}
\subsection{Splitting LLM}
LLMs are a pivotal development in the field of natural language processing (NLP), designed to learn linguistic patterns from extensive textual data~\cite{zhao2023llmsurvey}.
These models are proficient in performing an array of NLP tasks, encompassing text generation, comprehension, question-answering, and summarization, among others.
While LLM has high performance in text generation, it is computationally expensive and memory intensive due to the inherent structure of the LLMs and the immense number of parameters they encompass. 

\begin{figure*}[t!]
    \centering
    \includegraphics[width=\linewidth]{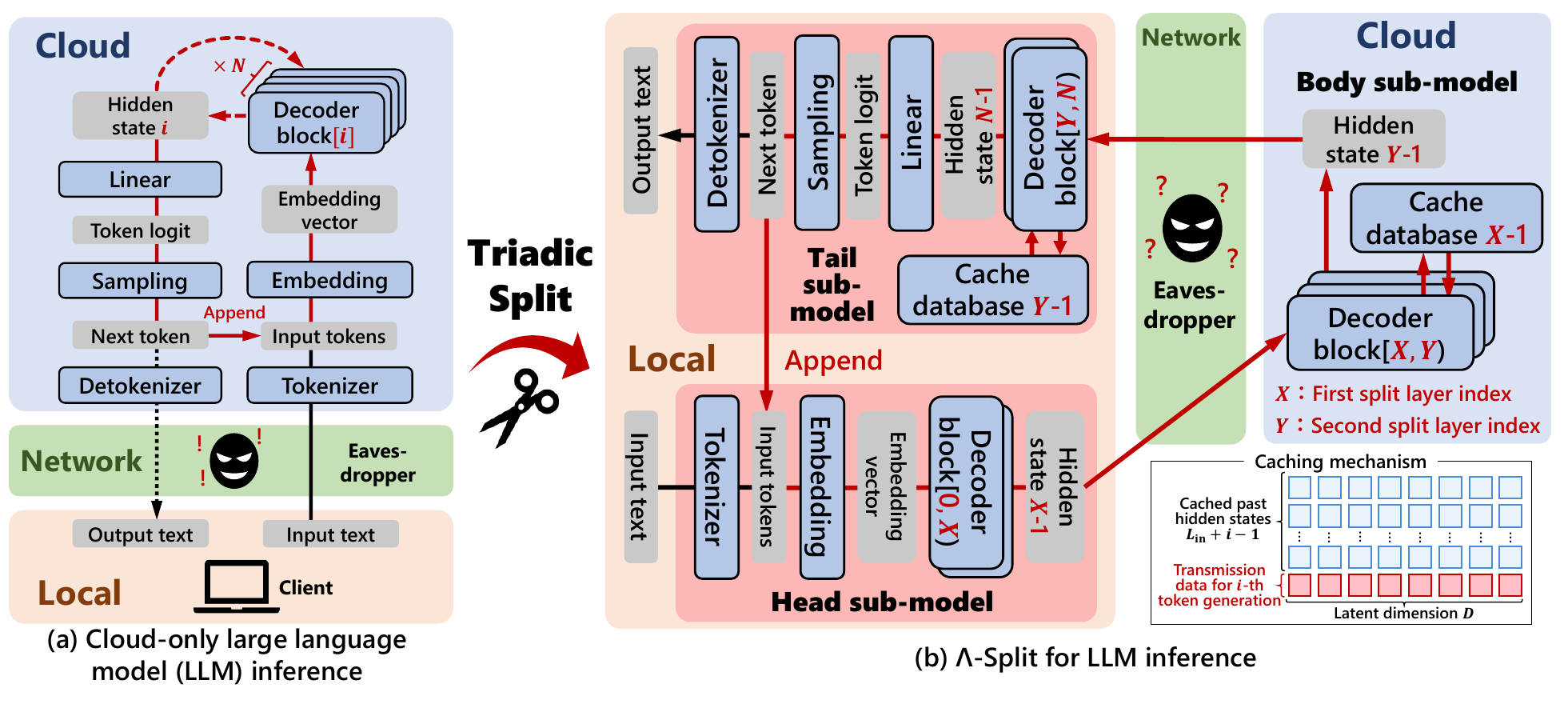}
    \caption{Conventional cloud-only and $\Lambda$-Split for LLM inference. 
    The $N$-layer stacked Transformer decoder blocks are split into three sub-models because the decoder blocks require large computations.
    LLM is split between $X-1$-th and $X$-th decoder layer, and $Y-1$-th and $Y$-th decoder layer.}
    \label{fig:framework_llm}
\end{figure*}

A preponderance of LLMs for text generation predominantly adopt an architecture that comprises a multi-layered stack of Transformer decoder blocks and generates text in an autoregressive fashion as delineated in \cref{fig:framework_llm}(a). 
In the conventional cloud-only LLM inference, as the input prompt from the user of the client device is received, it is transmitted to the cloud over the network.
In the cloud, the input prompt is tokenized and then transformed into embedding vectors that capture the semantic meaning of each token. 
These embedding vectors are subsequently fed into the $N$-layer stack of Transformer decoder blocks.
Subsequently, the model infers the next token, and the generated token is appended to the existing token sequence.
The LLM inference is repeated iteratively for the further next token generation until the end-of-sequence (EOS) token is generated or the sequence reaches a predetermined maximum length.
Finally, the generated token is converted back to text by the detokenizer and returned to the local client over networks.

Among numerous candidate points for splitting, we advocate for splitting the LLM at its most resource-intensive segment, specifically the deeply stacked Transformer decoder blocks, as illustrated in \cref{fig:framework_llm}(b).
$X$ and $Y$ indicate the first and second split layer indices, respectively.
$X$ and $Y$ are determined to be close to 0 and $N$, respectively, to increase the computational ratio in the cloud. 
The incorporation of the attention mechanism within the Transformer blocks, which enables each word in a sequence to establish relational computations with all other words, synergistically interacts with the multi-layered stacking to induce a quadratic escalation in both memory consumption and computational complexity as a function of sequence length.
Consequently, we can effectively offload the computational tasks to the cloud by splitting such that the majority of the Transformer blocks are encompassed within the body sub-model—allocated to the cloud server—and only a minimal number of Transformer blocks are included in the head and tail sub-models—allocated to the local device.
Additionally, only the hidden outputs of Transformer blocks are exchanged between the cloud and the local. 
These hidden outputs are mapped into a high-dimensional space, making them inherently difficult to interpret in isolation. 
As a result, even if intercepted, the likelihood of leaking privacy-sensitive information is substantially mitigated.

\subsection{Communication Traffic Reduction}
Simply transmitting hidden layer vectors of LLMs requires a very large bandwidth.
Specifically, the shape of hidden layer output for $i$-th token generation is $(L_\mathrm{in} + i, D)$, where $L_\mathrm{in}$ is the length of the input token sequence and $D$ is the dimension of embedding.
The value of $D$ is often set to a substantial value, such as 4096, 5120, or more, signifying that the vectors in the hidden layers possess considerable dimensionality.
Additionally, as mentioned above, LLMs for text generation are typically autoregressive models, i.e., they necessitate communication for each token generation when the model is split between cloud and local.
Hence, the order of total transmission data volume for generating text with token length $L_\mathrm{out}$ is $\mathcal{O}((L_\mathrm{in}+L_\mathrm{out}) L_\mathrm{out}D) = \mathcal{O}(L^2D)$, where $L$ is total token sequence length $L_\mathrm{in}+L_\mathrm{out}$, because the vector with the shape $(L_\mathrm{in} + i, D)$ is required to be transmitted at the $i$-th token generation.

To reduce the amount of transmitted data, we propose a simple caching mechanism.
We focus on the characteristics of autoregressive models; the hidden layer vector values of the past sequence, except the latest ones, remain unchanged. 
We cache the hidden layer vectors of the head sub-model in the cloud and those of the body sub-model in the local, as depicted in \cref{fig:framework_llm}(b). 
Thus, the transmitted vector for $i$-th token generation is only the latest hidden layer vector, whose shape is $(1, D)$, except initially transmitting the hidden layer vector for the input prompt with the token length $L_\mathrm{in}$.
Consequently, the order of the cumulative transmitted data volume is denoted as $\mathcal{O}((L_\mathrm{in}+L_\mathrm{out}) D) = \mathcal{O}(LD)$, which indicates a significant diminishment in the requisite data transmission bandwidth without affecting text generation results.

\subsection{Experimental Evaluation}
\input{table/llm}

\noindent\textbf{Setup: }
We evaluated the $\Lambda$-Split and caching mechanism using Llama-2-7b-chat-hf, which is a chat-specific model with seven billion parameters~\cite{touvron2023llama2}.
In the Llama-2-7b-chat-hf, the number of stacked Transformer decoder blocks $N$ is 32, and the hidden layer latent dimension $D$ is 4096.
We adjusted the computational load ratio (Local : Cloud) to (2 : 30), (8 : 24), and (16 : 16), specifically, $(X, Y)$ for model splitting was set to (1, 31), (4, 28), and (8, 24), respectively, to compare the changes in text generation latency due to computational load distribution. 
For comparison, we prepared the ratios of (0 : 32) and (32 : 0) as benchmarks, representing cloud-only and local-only inferences, respectively.

For the LLM's prompt, we employed question sentences with a fixed token length of $L_{\text{in}} = 14 $, as delineated in \cref{tab:llm_exp}. 
Corresponding examples of generated output texts, each having a token length of $L_{\text{out}} = 365$, are also presented in the same table. 
During the data collection for evaluation, we standardized the output token length at $L_{\text{out}} = 300$ to account for the variability in output text length across different inferences.

We used Jetson AGX ORIN 15W as the local device, and a GPU server, whose CPU and GPU are Intel(R) Core(TM) i9-10940X CPU and NVIDIA GeForce RTX 3090, respectively, as the cloud server.
The data type in inference is 16-bit floating-point number (FP16), and the transmission data is also FP16.
We implemented data communication between the local device and the cloud server via HTTP communication using FastAPI\footnote{\url{https://fastapi.tiangolo.com} (accessed on October 15, 2023)}, a Python web framework.
The bandwidth between the local and the cloud was set at 1000\,Mbit/s.
We experimentally obtained communication volume by measuring packet capture (PCAP) files captured by tcpdump\footnote{\url{https://www.tcpdump.org} (accessed on October 15, 2023)}.

\begin{figure*}[t!]
    \centering
    \includegraphics[width=\linewidth]{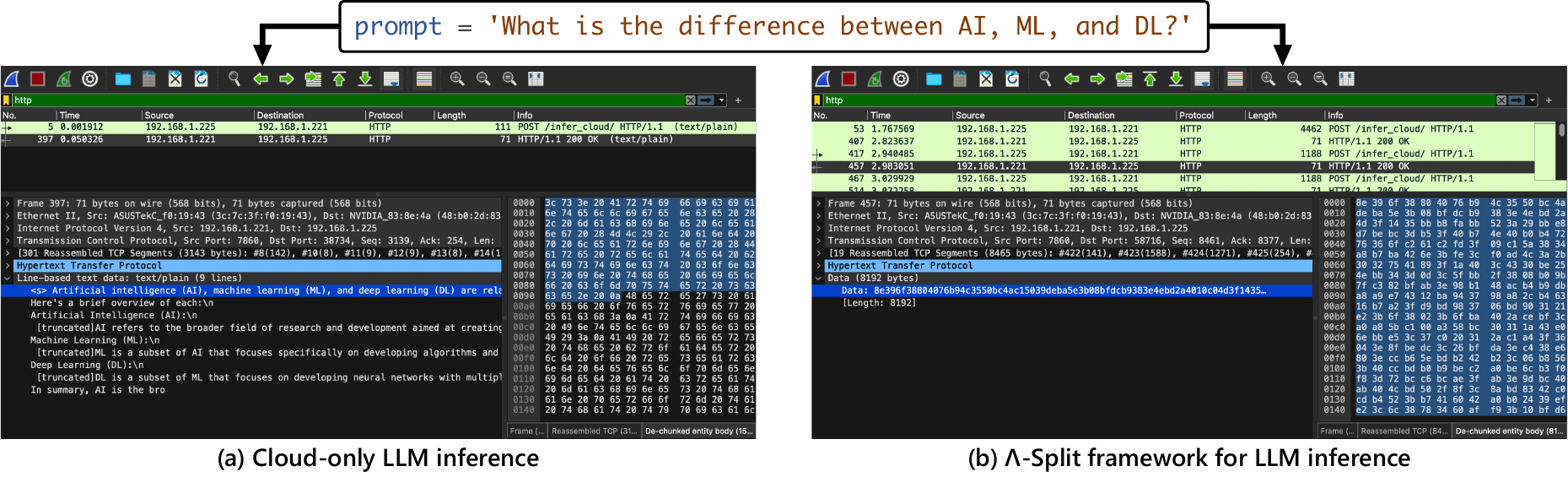}
    \caption{Visualization of example eavesdropped HTTP packets without encryption using Wireshark. The lower left windows show the decoded data in the HTTP packets transmitted from the cloud to the local. In the cloud-only inference, the input prompts and generated text can be known from the eavesdropped packets. Conversely, in $\Lambda$-Split, the decoded data consists of a byte stream representing the DNN's hidden output vector, thereby obfuscating the semantic content and complicating unauthorized interpretation.}
    \label{fig:pcap}
\end{figure*}

\noindent\textbf{Results: }
Initially, we consider a scenario in which an eavesdropper intercepts and demodulates communication packets, which are implemented using HTTP for this experiment. 
\cref{fig:pcap} delineates the outcomes of HTTP packet capture to elucidate the privacy-preserving capabilities conferred by the $\Lambda$-Split framework.
In the conventional cloud-only inference, an eavesdropper could readily decode both the prompt and the generated texts. 
In contrast, within the $\Lambda$-Split architecture, the data exposed to the eavesdropper are limited to the hidden states of DNN. 
Our empirical findings confirm the heightened difficulty an eavesdropper would encounter in decoding both the input and generated texts based solely on these intercepted hidden states.

\cref{tab:llm_exp} presents the empirical results for scenarios with the caching mechanism either enabled or disabled, as well as for various computational load ratios between local and cloud. 
The table captures key performance metrics, including the types of transmitted data, generated communication volume, latency, and average system throughput. 
While $\Lambda$-Split experiences a decline in generation throughput due to elevated latency compared to the cloud-only inference, $\Lambda$-Split significantly alleviates latency via computational offloading when compared with the local-only inference, which is the most secure option as it precludes external data exposure. 
Consequently, $\Lambda$-Split offers an inference framework that achieves higher generation throughput than the local-only inference while also providing superior privacy preservation compared with the cloud-only inference.

In $\Lambda$-Split, increasing the proportion of computations offloaded to the cloud, as exemplified in the (Local : Cloud) = (2 : 30) configuration, results in a decrease in total latency and an improvement in average token generation throughput. 
It should be noted that these results are predicated on experiments conducted with a single local device. 
In contrast, in scenarios involving multiple local devices, excessive offloading to the cloud could potentially lead to increased computational delays in the cloud infrastructure, thereby counterintuitively augmenting the overall system latency.

Furthermore, the proposed caching mechanism led to a substantial reduction in generated communication traffic, achieving a 99\% diminution. 
As a consequence, the total latency experienced a significant reduction, leading to an enhancement in the system throughput. 
This is attributable to the substantial reduction in the number of tokens transmitted between the cloud and local devices, facilitated by the caching mechanism.

\section{$\Lambda$-Split for Diffusion Model-based Image Generation}
\subsection{Splitting Diffusion Models}
Diffusion models, represented by denoising diffusion probabilistic models (DDPMs), are sophisticated DNN models utilized for image generation, texture synthesis, super-resolution, and modification tasks~\cite{yang2023dmsurvey}. 
The core concept of DDPM revolves around the probabilistic reconstruction of original images from their noise-perturbed counterparts, essentially learning the process of restoring original images from noisy observations, a technique referred to as denoising. 
The model employs a temporal addition of noise, transforming the image progressively until it converges to a state of pure noise, and subsequently learns to reverse this process to recover the original image. 
This innovative approach offers high quality and diversity over other generative models, such as generative adversarial networks (GANs) and variational autoencoders (VAEs).
Diffusion models continue to evolve, and one variation, the latent diffusion model (LDM), denoises against the latent vector of the image rather than the pixels.

\begin{figure*}[t!]
    \centering
    \includegraphics[width=\linewidth]{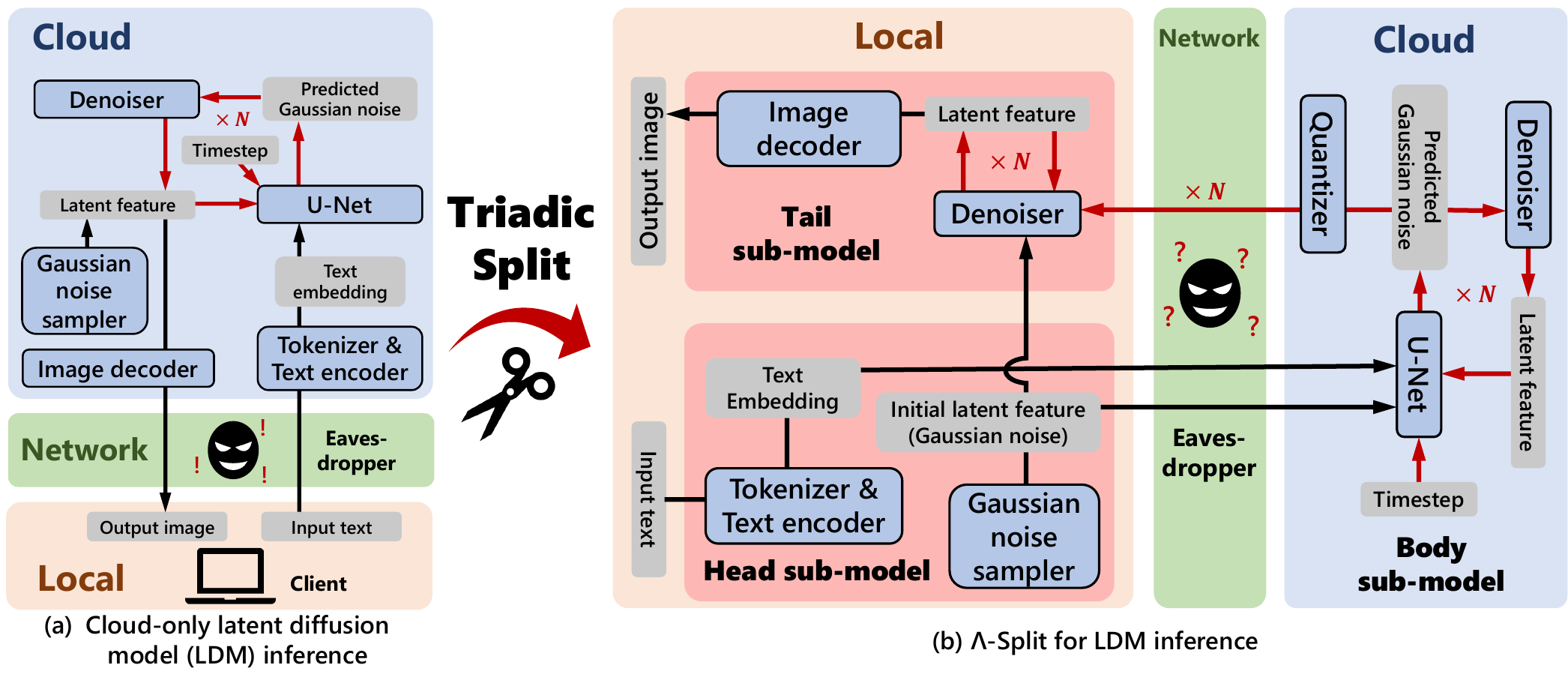}
    \caption{Conventional cloud-only and $\Lambda$-Split for LDM inference.
    LDM is triadically split so that the computationally expensive U-Net is located in the cloud and that most of the transmitted data is noise data.}
    \label{fig:framework_ldm}
\end{figure*}

\cref{fig:framework_ldm}(a) illustrates the conventional cloud-only text-to-image generation pipeline using LDM.
Firstly, the input prompt is transmitted from the local to the cloud, and text embeddings are inferred using a tokenizer and a text encoder. 
Subsequently, an initial latent feature sampled from Gaussian noise, the text embedding, and timestep information are input into the U-Net for noise prediction, yielding a predicted Gaussian noise. 
Based on the predicted noise, the latent feature of image is updated using the denoiser.
This process of noise prediction and latent feature updating is repeated multiple timesteps. 
Once all the timesteps are concluded, an image is decoded from the latent feature using an image decoder such as a VAE decoder, and the output image is transmitted from the cloud back to the local. 
However, in the conventional LDM inference pipeline, the input prompt and generated images themselves are transmitted over networks, which poses the risk of eavesdropping and data leakage.
Moreover, the U-Net used for noise prediction is frequently composed of Transformers and convolutions, requiring a substantial amount of computation. 
This computational requirement makes it challenging to infer solely on local devices with less computational power.

In the $\Lambda$-Split architecture, the LDM is strategically divided into three sub-models, as illustrated in \cref{fig:framework_ldm}(b). 
This triadic splitting aims to optimize the trade-offs among computational offloading, privacy preservation, and communication overhead. 
The head sub-model, deployed on the local device, integrates a tokenizer and text encoder for text embedding, along with a Gaussian noise sampler to initialize the latent feature. 
These generated text embeddings and initial latent feature are then transmitted to the cloud for further processing.
The body sub-model, hosted in the cloud, incorporates a U-Net for noise prediction and a denoiser to update the latent feature. 
Predicted Gaussian noise values are sent back to the local device at each timestep.
The tail sub-model, also situated locally, consists of a denoiser and an image decoder. The latent feature undergoes local updates based on the Gaussian noise received from the cloud at each timestep. 
The final output image is subsequently inferred from this updated latent feature using an image decoder, such as a VAE decoder.
In this splitting scheme, the data transmitted consists solely of the embedded texts and the initial and predicted Gaussian noise. 
This selective transmission renders it challenging to reconstruct the semantic content, thereby enhancing security against potential eavesdropping or data leakage.

\subsection{Communication Traffic Reduction}
While the $\Lambda$-Split-based LDM enables privacy-preserving distributed inference, the inherent nature of diffusion models, which typically require a multiple number of denoising steps $N$, can generate considerable communication traffic.
To mitigate traffic requirements, we employ quantization techniques to reduce the volume of transmitted data, particularly targeting the predicted noise values. 
Specifically, we employ affine quantization, as detailed in~\cite{wu2020integer}, which quantizes the values to integer representations in the body sub-model and dequantizes in the tail sub-model.
Given that the predicted noise adheres to a standard normal distribution, it is anticipated that 8-bit integer (INT8) quantization will achieve a favorable trade-off between communication volume and quantization error. 
The employment of INT8 quantization results in a 75\% reduction in the volume of data transmitted, compared to the use of 32-bit floating-point numbers (FP32).

\subsection{Experimental Evaluation}
\noindent\textbf{Setup: }
We evaluated $\Lambda$-Split for LDMs using stable-diffusion-xl-base-1.0, which is a variation of SDXL, an open LDM developed by Stability AI, and obtained transmitted data shapes, data volume, and generated images. 
We compared the generated images when the predicted noise was quantized from FP32 to FP16, INT8, INT6, INT4, and INT2 by affine quantization~\cite{wu2020integer}.
In addition to the qualitative evaluation, we quantitively evaluated the quantization methods using two image similarity metrics: peak-signal-to-noise ratio (PSNR) and structural similarity index measure (SSIM)~\cite{psnrssim}.
PSNR and SSIM indicate that the higher the value, the greater the similarity to the compared image.
For each experiment, we fixed the number of denoising steps $N$ to 50 and the random seed to 42.

\begin{figure*}[t!]
    \centering
    \includegraphics[width=\linewidth]{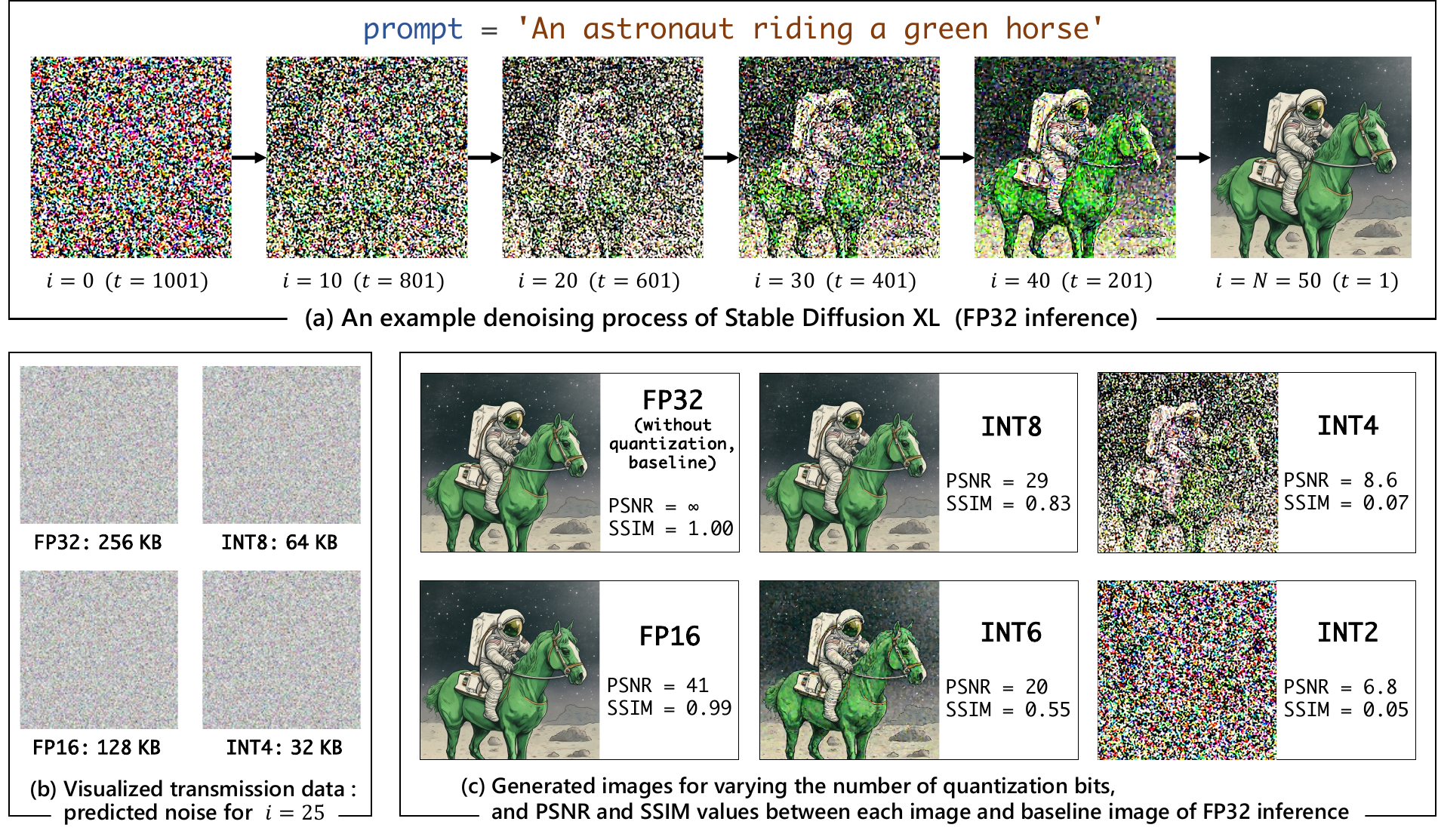}
    \caption{Example denoising process, transmission data, and generated images in $\Lambda$-Split for LDMs.
    The majority of transmission data is Gaussian noise.
    There was a tradeoff between traffic volume and quality of generated images.}
    \label{fig:evaluation_ldm}
\end{figure*}

\noindent\textbf{Results: }
\cref{fig:evaluation_ldm}(a) shows an example denoising process of SDXL without quantization.
As the denoising iteration progresses, the noise is removed and the final output is an image with content consistent with the input prompt.
\cref{fig:evaluation_ldm}(b) depicts the predicted noise for $i=25$\,th denoising iteration for four quantization methods, FP32, FP16, INT8, and INT4, as example transmission data.
Specifically, the Gaussian noise shape of the latent feature is (1, 4, 128, 128) in SDXL, which is converted to an RGBA image.
Regardless of the number of quantization bits, all images are noises, and it is considered difficult to recover the meaning of any image by itself.
Therefore, we consider that $\Lambda$-Split can enhance the privacy-preservation of transmitted data in distributed generative LDMs because most of the transmission data is the Gaussian-distributed predicted noise between the cloud and local devices.

\cref{fig:evaluation_ldm}(c) illustrates generated images and PSNR/SSIM values for varying the number of quantization bits.
Comparing the FP16 and INT8 images with the baseline FP32 image, it is possible to generate an image of such high quality that the difference is not noticeable to the naked eye.
In quantitative evaluation, the PSNR/SSIM values for FP16 and INT8 are 41/0.99 and 29/0.83, respectively, indicating a high similarity to the FP32 image. 
In contrast, as the number of quantization bits decreases as INT6, INT4, and INT2, more noise remains in generated images.
The PSNR and SSIM values of INT6, INT4, and INT2 have significantly decreased compared with FP16 and INT8, thus the similarity to the FP32 image is considered to be low.
Both quantitative and qualitative assessments indicate that the $\Lambda$-Split-based LDM, augmented with traffic reduction through quantization, can generate high-quality images with privacy preservation.

\section{Challenges and Potential Research Opportunities}
In this section, we discuss the challenges and potential research opportunities associated with $\Lambda$-Split, aimed at advancing more sustainable cloud-powered generative AI.

\vspace{2mm}
\noindent\textbf{Enhancing the Reliability of Privacy-Preservation Mechanisms: }
While $\Lambda$-Split-based generative AI aims to preserve privacy by transmitting hidden output vectors instead of the actual prompts or generated content, the extent to which this approach is secure remains theoretically unexplored. 
Specifically, the difficulty of inferring prompts or generated content from hidden output is strongly dependent on the model architecture and the layer at which the split occurs. 
Existing research in split computing has already reported attack methods that train 'inversion models' to infer inputs from hidden output vectors~\cite{dong2021deep}. 
Theoretical quantification of the security level against such attacks is a critical research challenge.

Further efforts to enhance reliability also constitute an important research agenda. 
The integration of existing techniques such as physical layer security~\cite{physicallayersecurity} and multi-party computation (MPC)~\cite{dong2023puma} appears to be a promising approach. 
Research into model architectures and training techniques that improve resistance to eavesdropping is also of significant interest. 
For instance, one could consider modifying the loss function during model training to incorporate the loss from an inversion model, treating it similarly to the discriminator in GANs. 
This approach would make the hidden output at the splitting point more resistant to inversion, and introduce a new performance metric of eavesdropping resistance into machine learning.

\vspace{2mm}
\noindent\textbf{Communication-Efficient Split Computing and Model Customization: }
While this study explores methods for reducing communication traffic, further reductions are essential for effective deployment in bandwidth-limited wireless networks. 
Model customization is a key strategy for achieving both communication and computational efficiency. 
In traditional SC, bottleneck architectures~\cite{matsubara:sc_survey} and neural architecture search (NAS)~\cite{nasc} have been employed to adapt communication and computation to local device capabilities. 
Techniques for tuning models to channel-specific characteristics, such as packet loss resilience, have also been considered~\cite{itahara_access}. 
The unique attributes of $\Lambda$-Split, including the presence of both head and tail sub-models on the local device, provide new avenues for research aimed at further reducing communication traffic and computational load.

In another way, optimizing the efficiency of model training for such customized models is a significant challenge. 
This is particularly true when considering constraints that prevent data from leaving the local device, even during training. 
Therefore, it is imperative to explore new model update methods for $\Lambda$-Split that combine efficient additional learning techniques such as low-rank adaptation (LoRA)~\cite{hu2022lora} with privacy-preserving learning methods such as FL~\cite{itahara_TMC} or split learning~\cite{split_learning}, particularly those that are communication-efficient.

\section{Conclusion}
In this article, we conceived $\Lambda$-Split, a split computing framework for generative AI leveraging the large computational power of the cloud server while preserving privacy.
To safeguard both input prompts and generated content from potential eavesdroppers across network and wireless channels, $\Lambda$-Split employs a triadic splitting of the generative AI model. 
In this configuration, the local device and the cloud engage solely in the exchange of the DNN's hidden output vectors over the network, thereby enhancing data security.
Our experimental results demonstrate that the $\Lambda$-Split can be applied to the state-of-the-art generative models and realize better tradeoffs between privacy preservation and generation speed than cloud-only computing and local-only computing.

\section*{Acknowledgment}
This work was supported by JST, PRESTO Grant Number JPMJPR2035, Japan.

\bibliographystyle{IEEEtran}
\bibliography{main}

\begin{IEEEbiographynophoto}{Shoki Ohta}~received the B.E.\ degree in information and communications engineering from Tokyo Institute of Technology in 2022.
He is currently studying toward the M.E.\ degree at the School of Engineering, Tokyo Institute of Technology.
He received the IEEE Vehicular Technology Society (VTS) Japan Young Researcher’s Encouragement Award and Outstanding Student Award from the Department of Information and Communications Engineering, Tokyo Institute of Technology, in 2022.
He is a graduate student member of IEEE.
His research interests include artificial intelligence-enabled next-generation wireless networks, and machine learning in wireless networks.
\end{IEEEbiographynophoto}

\begin{IEEEbiographynophoto}{Takayuki Nishio}~(S'11-M'14-SM'20) received the B.E.\ degree in electrical and electronic engineering and the master's and Ph.D.\ degrees in informatics from Kyoto University in 2010, 2012, and 2013, respectively. He had been an assistant professor in the Graduate School of Informatics, Kyoto University from 2013 to 2020. From 2016 to 2017, he was a visiting researcher in Wireless Information Network Laboratory (WINLAB), Rutgers University, United States. He has been an associate professor in the School of Engineering, Tokyo Institute of Technology, Japan, since 2020. His current research interests include machine learning-based network control, machine learning in wireless networks, and heterogeneous resource management.
\end{IEEEbiographynophoto}

\end{document}

%% file: table/llm.tex
\begin{table*}[t!]
\caption{Experimental settings and results for $L_\mathrm{out} = 300$ token generation with $L_\mathrm{in} = 14$ token input,\\and example input and output text of Llama 2 7B in $\Lambda$-Split.}
\label{tab:llm_exp}
\centering
\scalebox{0.95}{
\begin{tabular}{cc|c|ccc|cccc|c}
\toprule
\multirow{4}{*}{\begin{tabular}{c} Method\end{tabular}} & 
\multirow{4}{*}{\begin{tabular}{c} Computation\\layer ratio\\Local : Cloud\end{tabular}} & 
\multirow{4}{*}{\begin{tabular}{c} Transmission\\data\end{tabular}} & 
\multicolumn{3}{c|}{Communication volume (MB)} & \multicolumn{4}{c|}{Latency (s)} & 
\multirow{4}{*}{\begin{tabular}{c} Average\\generation\\throughput\\(token/s) \end{tabular}} \\ \cmidrule{4-10}
                            &            &  & \multirow{2.5}{*}{Uplink} & \multicolumn{1}{c|}{\multirow{2.5}{*}{Downlink}} & \multirow{2.5}{*}{Total} & \multicolumn{1}{c|}{\multirow{2.5}{*}{Communication}} & \multicolumn{2}{c|}{Computation}  & \multirow{2.5}{*}{Total} &                    \\ \cmidrule{8-9}
                            &            &  &                         & \multicolumn{1}{c|}{}                          &                        & \multicolumn{1}{c|}{}                               & Local & \multicolumn{1}{c|}{Cloud} &                        &                    \\ \midrule
Cloud-only        & \, 0 : 32 &  Text & 0.01                        & \multicolumn{1}{c|}{0.03}                          & 0.03                       & \multicolumn{1}{c|}{0.05}                               & 0.00     & \multicolumn{1}{c|}{20.46}      & 20.51                       & 14.62                   \\ \midrule
\multirow{3}{*}{\begin{tabular}{c} $\Lambda$-Split\\ \textbf{w/o} caching \\mechanism\end{tabular}}   & \, 2 : 30    &  \multirow{3}{*}{\begin{tabular}{c} Hidden\\state\end{tabular}} & 404.44                        & \multicolumn{1}{c|}{470.69}                          & 875.13                       & \multicolumn{1}{c|}{119.19}                               & 31.24     & \multicolumn{1}{c|}{19.74}      & 170.17                       & 1.76                   \\
                            & \, 8 : 24    &  & 403.74                        & \multicolumn{1}{c|}{470.15}                          & 873.90                       & \multicolumn{1}{c|}{119.87}                               & 82.38     & \multicolumn{1}{c|}{16.62}      & 218.87                       & 1.37                   \\
                            & 16 : 16    &  & 404.14                        & \multicolumn{1}{c|}{471.06}                          & 875.20                       & \multicolumn{1}{c|}{121.16}                               & 156.15     & \multicolumn{1}{c|}{12.11}      & 289.42                       & 1.03                   \\ \midrule
\multirow{3}{*}{\begin{tabular}{c} $\Lambda$-Split\\\textbf{w/} caching\\mechanism\end{tabular}}    & \, 2 : 30    &  \multirow{3}{*}{\begin{tabular}{c} Latest part of\\hidden state\end{tabular}} & 3.10                        & \multicolumn{1}{c|}{3.25}                          & \textbf{6.34}                       & \multicolumn{1}{c|}{2.72}                               & 26.94     & \multicolumn{1}{c|}{19.71}      &  49.37                      & \textbf{6.07}                   \\
                            & \, 8 : 24    &  & 3.10                        & \multicolumn{1}{c|}{3.26}                          & \textbf{6.36}                       & \multicolumn{1}{c|}{2.81}                               & 81.72     & \multicolumn{1}{c|}{16.11}      & 100.64                      & \textbf{2.98}                   \\
                            & 16 : 16    &  & 3.10                        & \multicolumn{1}{c|}{3.25}                          & \textbf{6.34}                       & \multicolumn{1}{c|}{2.89}                               & 157.51     & \multicolumn{1}{c|}{11.78}      & 172.18                       & \textbf{1.74}                   \\ \midrule
Local-only        & 32 : \,\,\,0  &  $-$ & 0.00                        & \multicolumn{1}{c|}{0.00}                          & 0.00                       & \multicolumn{1}{c|}{0.00}                               & 300.77     & \multicolumn{1}{c|}{0.00}      & 300.77                       & 0.99                   \\ \bottomrule
\toprule
Input text & \multicolumn{10}{l}{\begin{tabular}{p{160mm}}{\scriptsize What is the difference between AI, ML and DL? ($L_\mathrm{in} = 14$ tokens)}\end{tabular}} \\ \midrule
Output text & \multicolumn{10}{l}{\begin{tabular}{p{160mm}}{\scriptsize Artificial intelligence (AI), machine learning (ML), and deep learning (DL) are related but distinct concepts in the field of computer science. 
Here's a brief overview of each:
Artificial Intelligence (AI):
AI refers to the broader field of research and development aimed at creating machines that can perform tasks that typically require human intelligence, such as understanding language, recognizing images, making decisions, and solving problems. AI involves a wide range of techniques, including rule-based systems, decision trees, and expert systems.
Machine Learning (ML):
ML is a subset of AI that focuses specifically on developing algorithms and statistical models that enable machines to learn from data, without being explicitly programmed. ML involves training machines to recognize patterns in data, make predictions or decisions, and improve their performance over time. Common ML techniques include supervised and unsupervised learning, reinforcement learning, and deep learning.
Deep Learning (DL):
DL is a subset of ML that focuses on developing neural networks with multiple layers, inspired by the structure and function of the human brain. DL algorithms are capable of learning and improving on their own by automatically adjusting the connections between layers, allowing them to learn and represent complex patterns in data. DL has been particularly successful in areas such as computer vision, natural language processing, and speech recognition.
In summary, AI is the broader field that encompasses both ML and DL, while ML is a subset of AI that focuses on developing algorithms that enable machines to learn from data, and DL is a subset of ML that focuses on developing neural networks with multiple layers to learn and represent complex patterns in data.
    ($L_\mathrm{out} = 365$ tokens)}\end{tabular}} \\ \bottomrule
\end{tabular}
}
\end{table*}